\documentclass[11pt]{article}

\usepackage[final]{acl}

\usepackage{times}
\usepackage{latexsym}

\usepackage[T1]{fontenc}

\usepackage[utf8]{inputenc}

\usepackage{microtype}

\usepackage{inconsolata}

\usepackage{graphicx}
\usepackage{enumitem}
\usepackage{amsmath}
\usepackage{multirow}
\usepackage{booktabs}
\usepackage[table]{xcolor} 
\usepackage{cleveref}
\usepackage{xspace} 
\usepackage{calc}
\usepackage{siunitx}
\usepackage{subcaption}
\usepackage[T1]{fontenc}
\usepackage{tabularx}
\usepackage{pifont}
\usepackage{fontawesome5}
\usepackage{hyperref}

\newcommand{\framework}{\textsc{Deco-G}\xspace}

\DeclareMathOperator{\topk}{top-k}

\title{Decoupling Task-Solving and Output Formatting in LLM Generation}

\author{
Haikang Deng, Po-Nien Kung, Nanyun Peng\\
University of California, Los Angeles\\
\texttt{\{haikang, ponienkung, violetpeng\}@cs.ucla.edu}
}

\begin{document}
\maketitle
\begin{abstract}

Large language models (LLMs) are increasingly adept at solving complex problems, such as mathematical reasoning and automatic evaluation. However, performance often degrades when prompts intertwine task instructions with rigid formatting requirements. This entanglement creates competing goals for the model, hindering its reasoning capabilities. 
To address this, we introduce \framework, a decoding framework that explicitly decouples format adherence from problem solving. \framework delegates format adherence to a separate Format Estimation Module (FEM), which performs probabilistic lookahead to estimate future format compliance rate and reweighs token probabilities, allowing the LLM to focus solely on task resolution. To make this approach both practical and efficient, we introduce three key innovations: instruction-aware distillation, a flexible trie-building algorithm, and HMM state pruning. Experiments across mathematical reasoning, event argument extraction, and LLM-as-a-judge demonstrate that \framework constantly gains over prompting or structured generation baselines, with guaranteed format compliance. We release our code at
\href{https://github.com/haikangdeng/deco-g}
{\faGithub\ \texttt{haikangdeng/deco-g}}.

\end{abstract}
\section{Introduction}

\begin{figure*}[htbp]
    \centering
    \includegraphics[width=0.95\linewidth, trim={0.6cm, 0.8cm, 0.6cm, 0.2cm}, clip]{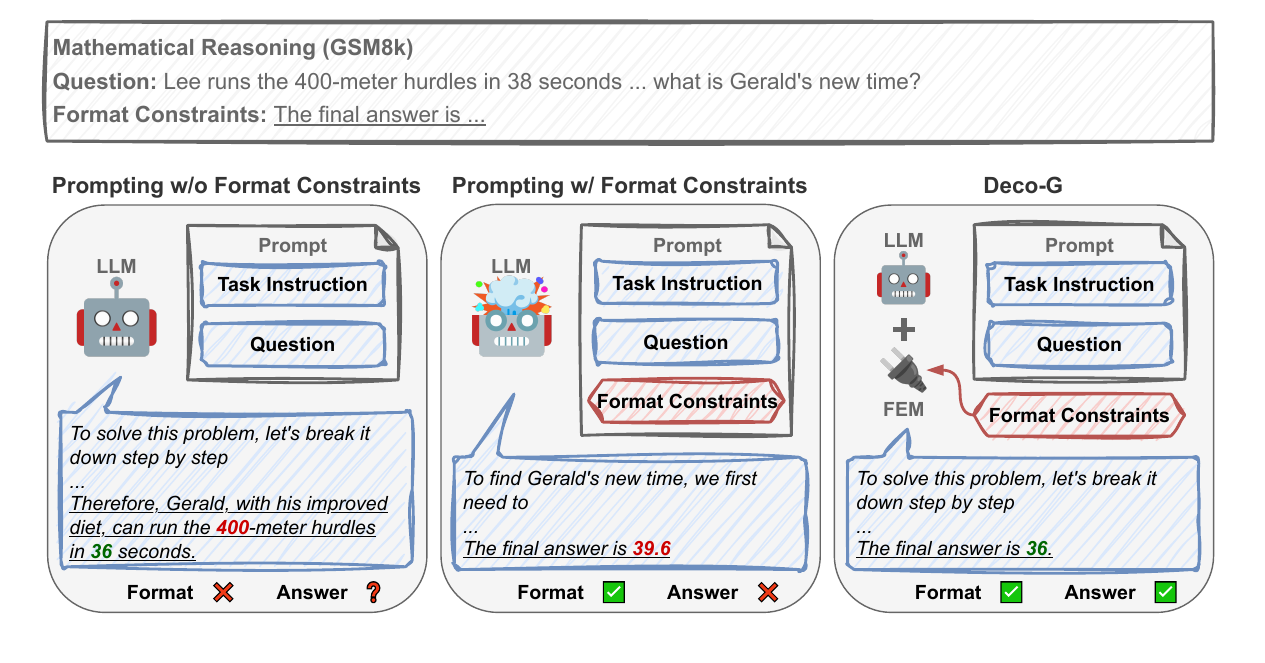}
    \caption{GSM8k examples. Unconstrained prompting yields correct answers but fails formatting, while adding constraints degrades reasoning accuracy. \framework decouples formatting via a Format Estimation Module (FEM), achieving both strict compliance and correct answer.}
    \vspace{-0.6em}
    \label{fig:teaser}
\end{figure*}

Instruction fine-tuning~\citep{wei2021finetuned, chung2024scaling} empowers LLMs to solve complex tasks, often enhanced by reasoning strategies like Chain-of-Thought~\citep{wei2022chain} and Tree-of-Thought~\citep{yao2023tree}. However, emerging evidence suggests that combining problem-solving instructions with strict output-format requirements in a single prompt negatively impacts performance~\citep{lmsf, long-etal-2025-llms, he2024doespromptformattingimpact}. For instance, \citet{long-etal-2025-llms} demonstrate that output structure significantly affects accuracy on benchmarks like MMLU~\citep{hendrycks2020measuring}, while \citet{lmsf} observe that stricter constraints correlate with greater reasoning degradation. This suggests that the current paradigm of intertwining task and format instructions (as shown in \Cref{fig:teaser}) may hurt LLMs' reasoning capabilities.


To mitigate this, recent works have explored relaxing format strictness~\citep{lmsf} or adopting more intuitive schema~\citep{long-etal-2025-llms, he2024doespromptformattingimpact}. Yet, these adjustments still impose constraints that distract LLMs from reasoning. Alternatively, constrained decoding frameworks~\citep{pmlr-v235-beurer-kellner24a, guidance_ai_2024_guidance, outlines} ensure compliance by strictly enforcing token transitions. However, this rigid intervention does not consider the model's internal reasoning flow, resulting in abrupt cut-offs or incoherent outputs. This trade-off highlights the critical need for a framework that seamlessly decouples format constraints from task solving to unlock the full potential of LLMs.


In this paper, we introduce \framework, a framework that explicitly decouples format adherence from task reasoning, allowing LLMs to focus solely on problem-solving while introducing an auxiliary module to guarantee format adherence. 
Specifically, it leverages the modularity of existing controllable text generation methods (e.g. GeLaTo~\citep{gelato}, Ctrl-G~\citep{ctrlg}) and delegates format adherence to a dedicated Format Estimation Module (FEM) that continuously estimates the final format compliance and reweighs each next-token distribution to guide generation.

While this work builds upon GeLaTo~\citep{gelato} and Ctrl-G~\citep{ctrlg}, it is, to our knowledge, the first framework that introduces explicit separation of task solving and format adherence to preserve LLMs' full potential. 
Moreover, naively applying existing methods leads to degraded performance, due to their limited compatibility with instruction-tuned LLMs and computational bottlenecks when handling complex templates. 
To this end, we introduce three innovations: (1) instruction-aware distillation to capture task-oriented behaviors; (2) a flexible trie algorithm for efficient automata construction; and (3) HMM state pruning to accelerate inference. 

To assess \framework's effectiveness across diverse constraint types, we evaluate it on three distinct tasks: mathematical reasoning, event argument extraction, and LLM-as-a-judge. Experiments demonstrate that \framework consistently improves overall task performance, driven by: (1) guaranteeing 100\% format compliance; (2) enabling more natural, context-aware format integration; and (3) freeing the LLM to concentrate on reasoning without the cognitive burden of formatting.

Our contributions are as follows:
\setlist{nolistsep}
\begin{itemize}[topsep=0.1em, itemsep=0.1em, leftmargin=16pt]
    \item We propose a novel decoding framework that uses a Format Estimation Module to explicitly decouple format adherence from task solving, preserving the LLM's reasoning capabilities while enforcing strict constraints.
    \item We introduce three technical innovations---instruction-aware distillation, flexible trie construction, and HMM state pruning---that induce minimal computational overhead and render the \framework framework practical for real-world deployment.
    \item We demonstrate consistent performance gains across diverse benchmarks, supported by a detailed analysis of the steering mechanism and entropy-based control dynamics.
\end{itemize}

\section{Preliminaries}
This section reviews prior work on controllable generation that our approach builds upon, and highlights how these methods, while instrumental, struggle to achieve our goal of efficient task–format disentanglement. Our formulation and technical contributions are introduced in the next section.

\subsection{Generation with Attribute Control}
We follow prior work to formulate controllable text generation given a desired attribute $\alpha$ as
\begin{equation*}
P(x_{1:n}|\alpha) = \prod_{t}{P(x_t | x_{<t}, \alpha)},
\end{equation*}
where the objective is to optimize the probability of sequences that exhibits the attribute $\alpha$ (e.g., contain certain keywords, sentiment, etc.). 
At each generation step $t$, the target distribution for producing text with the desired attribute is $P(x_t | x_{<t}, \alpha)$. Using Bayes' rule, we can rewrite this as:
\begin{equation}
    P(x_t | x_{<t}, \alpha) = P_{\mathrm{LM}}(x_t | x_{<t}) \frac{P_{\mathrm{LM}}(\alpha | x_{t}, x_{<t})}{P_{\mathrm{LM}}(\alpha | x_{<t})}
    \label{eq: posterior}
\end{equation}
The first term $P_{\mathrm{LM}}(x_t | x_{<t})$ is the language model's next-token probability. The second term $\frac{P_{\mathrm{LM}}(\alpha | x_t, x_{<t})}{P_{\mathrm{LM}}(\alpha | x_{<t})}$ acts as a control signal. It quantifies how the choice of the current token $x_t$ influences the probability that the complete sequence will satisfy attribute $\alpha$.
However, directly calculating this ratio is intractable, as it requires marginalizing over all possible future sequences. A key challenge in controllable generation is to find a tractable approximation for this term.

\subsection{Estimating Attribute Likelihood}
Recent controllable generation frameworks such as GeLaTo~\citep{gelato} and Ctrl-G~\citep{ctrlg} leverage a tractable probabilistic model (TPM) to efficiently estimate the marginal probability $P(\alpha | x_t, x_{<t})$, serving as a signal to steer an LLM's generation, following
\begin{equation}
P(x_t | x_{<t}, \alpha) \propto P_{\mathrm{LM}}(x_t | x_{<t}) P_{\mathrm{TPM}}(\alpha | x_t, x_{<t})
\label{eq:steering}
\end{equation}
These approaches first distill a Hidden Markov Model (HMM) as a probabilistic approximation of the LLM and then encode logical constraints to formal structure.

\paragraph{Sequence Modeling.}
HMM is the specific type of TPM used in these frameworks, chosen for its ability to model sequential data tractably. The joint probability distribution over a sequence of observed variables (tokens, $x_{1:n}$) and a corresponding sequence of hidden state variables $z_{1:n}$, is modeled as 
\begin{equation*}
\begin{split}
& P_{\mathrm{HMM}}(x_{\leq t}, z_{\leq t}) = \\
& \quad P(z_1)P(x_1|z_1) \prod_{t=2}^{T} P(z_t|z_{t-1})P(x_t|z_t)
\end{split}
\end{equation*}
HMM's Markov property enables efficient probabilistic inference over future sequences, a task that is intractable for language models. In GeLaTo and Ctrl-G, the HMM is distilled from the LLM using samples drawn unconditionally from the LLM. 

\paragraph{Formalizing Constraints.}
To enforce a constraint using the HMM, the constraints must be expressed in a formal language. \citet{ctrlg} propose representing logical constraints as Deterministic Finite Automata (DFA). A DFA is an abstract state machine that recognizes patterns in sequences. Formally, a DFA is a 5-tuple $\mathcal{D} = (Q, \Sigma, \delta, q_0, F)$, where $Q$ is a finite set of states, $\Sigma$ is the alphabet (the LLM's token vocabulary), $\delta: Q \times \Sigma \to Q$ is the transition function, $q_0 \in Q$ is the initial state, and $F \subseteq Q$ is the set of accept states. 
A sequence is ``accepted'' if it drives the machine from its initial state to an accept state. This formalism is capable of representing logical constraints including the presence of keyphrases and word counts.

\paragraph{Probabilistic reasoning over logical constraints.}
The core idea of these prior frameworks is to use the TPM to perform a \textit{probabilistic lookahead} — i.e., to efficiently compute $P_{\mathrm{TPM}}(\alpha | x_t, x_{<t})$, the probability that the full generated sequence will satisfy the constraint $\alpha$. This is accomplished by marginalizing the joint HMM-DFA state space over all possible future sequences that reach an accepting state in the DFA. According to \citet{ctrlg}, this marginalization can be calculated efficiently using a backward recurrence relation. Refer to \Cref{app:probabilistic} for detailed derivation.

\subsection{From Prior Work to \framework}
While GeLaTo~\citep{gelato} and Ctrl-G~\citep{ctrlg} demonstrate the effectiveness of generic TPMs for logical constraints (e.g., keyphrases), they face significant limitations when adapted for task-format decoupling in instruction-tuned LLMs:
\begin{itemize}[topsep=0.1em, itemsep=0.1em, leftmargin=15pt]
    \item \textbf{Domain Shift:} Prior methods distill HMMs from unconditional random generations. This fails to capture the \textit{instruction-conditional} distribution required for tasks.
    \item \textbf{Template Complexity:} Real-world formats often interweave fixed text with variable-length fields. Representing these structures via standard DFA construction leads to state-space explosion, creating a computational bottleneck.
    \item \textbf{Inference Latency:} The computational cost of probabilistic lookahead scales poorly with the massive vocabulary size. Without optimization, this brings significant latency for real-time generation.
\end{itemize}

\section{\framework}
\begin{figure}
    \centering
    \includegraphics[width=1\linewidth, trim={0cm 0cm 0cm 0.2cm}, clip]{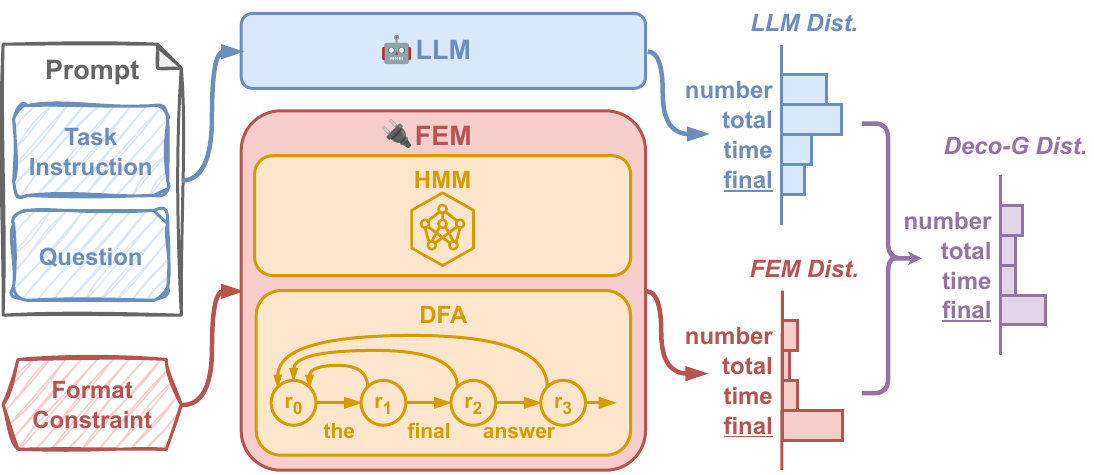}
    \caption{\framework decouples task and format by prompting LLM with task-only information and sending format constraints to FEM. \framework decodes from the posterior constructed by reweighing LLM token probabilities with the FEM estimated satisfaction rate.
    }
    \vspace{-1em}
    \label{fig:framework}
\end{figure}

In this section, we present \framework, a framework that realizes the decoupling of task reasoning from output formatting. As shown in \Cref{fig:framework}, our method separates the input prompt: the LLM receives only the task-specific information, while a dedicated Format Estimation Module (FEM) receives the format constraints. At each decoding step, the FEM estimates the likelihood of future compliance with the given format constraints $\alpha$. This likelihood is then used to reweigh the LLM's original token probabilities, steering the generation towards a format-compliant output. Below, we describe the key components that enable this framework.

\subsection{Instruction-Aware HMM Distillation}
An HMM can approximate a large language model's (LLM) output distribution to guide controllable generation. The fidelity of this approximation is critical---ideally, an HMM that perfectly replicates the LLM's probabilities would yield an exact posterior for format-decoupled generation, per Equation \eqref{eq: posterior}. Our key insight is that for instruction-tuned LLMs, the output distribution is fundamentally different when conditioned on a prompt versus when generating text unconditionally.
Prior methods~\citep{gelato, ctrlg}, however, distill their HMMs using text sampled unconditionally from a model, an approach that fails to capture the task-oriented behavior that emerges after instruction fine-tuning. This design renders methods like Ctrl-G ineffective, leading to the suboptimal control patterns we demonstrate in \Cref{app:unconditioned}.

To bridge this gap, we carry out instruction-aware distillation: conditioning an HMM on task-oriented behavior by training it exclusively on the LLM's instruction-response pairs. Specifically, we distill knowledge from over \textit{one million} completions generated by an LLM prompted with \textit{one thousand} unique instructions from the Natural-Instructions-v2~\citep{naturalinstructions} dataset. Following~\citet{gelato}, we train the HMM using the Baum-Welch algorithm~\citep{baum1972inequality}. This process yields a robust HMM that models the LLM’s conditional, instruction-following behavior, enabling more precise control over generation across a wide spectrum of tasks.

\subsection{Flexible Trie Building for Complex Constraints}
\label{sec:trie}

To efficiently enforce complex formatting requirements, we formalize the constraint space as a \textit{Template Language} $\mathcal{L}_{\mathcal{T}}$. Unlike standard trie constructions which branch on individual characters, our flexible trie algorithm constructs a Deterministic Finite Automaton (DFA) by branching on \textit{segments}, comprised of fixed Pivots and variable Wildcards.

\paragraph{Template Segments.}
Let $\mathcal{V}$ be the vocabulary of the LLM. A template $T \in \mathcal{T}$ is defined as an ordered sequence of segments $S = (s_1, s_2, \dots, s_n)$, where each segment $s_i$ is either:
\begin{itemize}[topsep=0.3em, itemsep=0em, leftmargin=15pt]
    \item \textbf{Pivot ($P$):} A deterministic sequence of tokens $x_{1:m} \in \mathcal{V}^m$ representing static text (e.g., formatting cues like ``\texttt{The answer is:}''). The language accepted by a pivot is the singleton set:
    \begin{equation*}
        L(P) = \{x_{1:m}\}
    \end{equation*}
    \item \textbf{Wildcard ($W$):} A variable slot constrained only by length $[\ell_{min}, \ell_{max}]$. It accepts any token sequence $x$ where the length satisfies the bounds. Its language is the union of all vocabulary permutations within the length range:
    \vspace{-0.5em}
    \begin{equation*}
        L(W) = \bigcup_{k=\ell_{min}}^{\ell_{max}} \mathcal{V}^k
    \end{equation*}
\end{itemize}
The language recognized by a full template $T$ is the concatenation of its segment languages: $L(T) = L(s_1) \cdot L(s_2) \cdots L(s_n)$.

\paragraph{Trie Algorithm.}
We construct a minimal DFA $\mathcal{A} = (Q, \Sigma, \delta, q_{init}, F)$ that recognizes the union of all allowed templates. The core innovation is our State Convergence strategy, which prevents state-space explosion for multi-part templates.
Let $q_{init}^{(i)}$ be the unique entry state for segment $s_i$. We construct the transition function such that for any valid string realization $w$ of the current segment $s_i$, the automaton transitions to the unique entry state of the next segment $s_{i+1}$:
\begin{equation}
    \forall w \in L(s_i), \quad \delta^*(q_{init}^{(i)}, w) = q_{init}^{(i+1)}
    \label{eq:convergence}
\end{equation}
where $\delta^*: Q \times \mathcal{V}^* \to Q$ is the extended transition function processing a string of tokens. 
Equation~\ref{eq:convergence} mathematically guarantees that the state configuration for segment $s_{i+1}$ is strictly independent of the path taken through $s_i$.

\paragraph{Complexity.} 
Standard DFA construction treats every valid length permutation of a wildcard as a distinct path. For a template with $N$ sequential wildcards of length variance $K$, this dependency chains forward, resulting in exponential complexity $O(K^N)$.
Consider the event argument extraction template alternating between fixed pivots $P$ and variable wildcards $W$:
\begin{equation*}
    T = P_1 (\text{``\texttt{Role A:}''}) \cdot W_1 \cdot P_2 (\text{``\texttt{Role B:}''}) \cdot W_2 \dots
\end{equation*}
In standard construction, the state history of $W_2$ depends on the specific length chosen for $W_1$. In contrast, our algorithm enforces the convergence property (Equation~\ref{eq:convergence}): all valid paths for $W_1$ merge back to the single canonical node $q_{init}^{(P_2)}$. This decouples the wildcard histories, ensuring the total state space grows linearly as $O(N \times K)$.

\subsection{Estimating Format Compliance}
With a distilled HMM that simulates LLM distribution and a DFA that encodes format constraints $\alpha$, we calculate the marginal probability over all sequences accepted by $\mathcal{D(\alpha)}$ as
\begin{equation*}
P_\mathrm{FEM}(\alpha|x_t,x_{<t})=\frac{P(\mathcal{D}(\alpha)=1, x_t, x_{<t})}{P(x_t,x_{<t})}
\end{equation*}
While the joint probability of format compliance and context sequence $P(\mathcal{D}(\alpha)=1, x_t, x_{<t})$ is not readily available in the FEM, we follow~\citet{ctrlg}'s marginalization of HMM over DFA (\Cref{app:probabilistic}) to calculate this value. Finally, we use the FEM estimated compliance rate as likelihood to construct the \framework posterior:
\begin{equation*}
P(x_t | x_{<t}, \alpha) \propto P_{\mathrm{LM}}(x_t | x_{<t}) [P_{\mathrm{FEM}}(\alpha | x_{<t}, x_t)]^\gamma
\end{equation*}
The resulting scores are re-normalized over the vocabulary $\mathcal{V}$ (via softmax) to ensure a valid probability distribution. $\gamma$ is a hyperparameter that controls the steering strength with a default value of $1$.

\subsection{HMM Hidden State Pruning}
\label{sec:pruning_methodology}

The Format Estimation Module (FEM) provides effective guidance but introduces a computational bottleneck during inference. The primary cost lies in the HMM's emission stage, which calculates the probability distribution over the entire vocabulary $\mathcal{V}$ from $h$ hidden states. With $h=4096$ and $|\mathcal{V}| > 100k$ for \textit{Llama} and \textit{Qwen}, this matrix-vector multiplication ($O(h|\mathcal{V}|)$) significantly impedes latency.

To mitigate this, we introduce \textbf{HMM hidden state pruning}. This technique exploits the observation that the probability mass of the hidden state distribution is highly concentrated within a small subset of states (see \Cref{fig:retention}). Instead of using the full state space, we approximate the emission probabilities by considering only the $\topk$ most probable states. Empirically, selecting just the top $5\%$ ($k=200$) is sufficient to retain over $98\%$ of the full model's performance. This strategy drastically improves efficiency by reducing the emission complexity from $O(h |\mathcal{V}|)$ to $O(k|\mathcal{V}| + h \log h)$, where the $O(h \log h)$ term accounts for selecting the top states. With $k \ll h$, this optimization achieves a $\sim 13\times$ reduction in FLOPs per decoding step, ensuring that the guidance overhead remains negligible while maintaining robust format compliance.

\section{Experiment}
\label{sec:experiment}

\paragraph{Experimental Setup.}
We assess \framework's overall performance over three tasks: (1) math problem solving with reasoning, (2) event argument extraction as a generative task, and (3) LLM-as-a-judge for summary evaluation (see \Cref{app:llm-as-a-judge}). We apply \framework on performant instruction models \textit{Llama-3.1-8B-Instruct}~\citep{grattafiori2024llama}, \textit{Qwen2.5-7B-Instruct}~\citep{qwen2025qwen25technicalreport}, and \textit{Qwen3-8B}~\citep{qwen3} to verify its effectiveness. The baselines we include for comparison are as follows:
\begin{itemize}[topsep=0.1em, itemsep=0.1em, leftmargin=16pt]
    \item \textbf{Prompt-Only (NL/JSON):} Standard greedy decoding conditioned on task instructions and output constraints (natural language or JSON), without external interference.
    \item \textbf{Structured Generation (NL/JSON):} Constrained decoding using \textit{Outlines}~\citep{outlines} to strictly enforce the target output format, either via a natural language template or a JSON schema.
    \item \textbf{Ctrl-G:} A controllable generation baseline that uses an HMM distilled from unconditional sampling to guide generation, as detailed in~\citet{ctrlg}.
\end{itemize}
For the following experiments, we adopt greedy decoding to ensure fair comparison with baseline methods and evaluate zero-shot performance.

\definecolor{modelrowcolor}{gray}{0.90}

\begin{table}[htbp]
\centering
\small
\setlength{\tabcolsep}{1pt}
\renewcommand{\arraystretch}{1.1}
\begin{tabular}{l l c c}
\toprule
\textbf{Model} & \textbf{Method} & \textbf{Format (\%)} & \textbf{Acc. (\%)} \\
\midrule
\multirow{6}{*}{\shortstack[l]{\textit{Llama-3.1-8B}\\\textit{-Instruct}}} 
 & Prompt-Only (NL) & 96.3 & 82.3 \\
 & Prompt-Only (JSON) & 64.7 & 51.8 \\
 & Outlines (NL) & 100 & 81.3 \\
 & Outlines (JSON) & 100 & 75.7 \\
 & Ctrl-G & 100 & 61.6 \\
 & \textbf{\framework} & 100 & \textbf{85.2} \\
\midrule
\multirow{6}{*}{\shortstack[l]{\textit{Qwen2.5-7B}\\\textit{-Instruct}}} 
 & Prompt-Only (NL) & 98.0 & 83.6 \\
 & Prompt-Only (JSON) & 93.3 & 74.8 \\
 & Outlines (NL) & 99.9 & 82.7 \\
 & Outlines (JSON) & 99.8 & 79.0 \\
 & Ctrl-G & 100 & 84.9 \\
 & \textbf{\framework}$^{\gamma=2}$ & 100 & \textbf{88.6} \\
\midrule
\multirow{6}{*}{\textit{Qwen3-8B}} 
 & Prompt-Only (NL) & 97.4 & 90.5 \\
 & Prompt-Only (JSON) & 66.9 & 61.4 \\
 & Outlines (NL) & 100 & 88.3 \\
 & Outlines (JSON) & 99.2 & 90.6 \\
 & Ctrl-G & 100 & 88.7 \\
 & \textbf{\framework}$^{\gamma=2}$ & 100 & \textbf{91.7} \\
\bottomrule
\end{tabular}
\caption{GSM8k Results. \framework guarantees format satisfaction and consistently outperforms Prompt-Only and Outlines baselines across all models.}
\vspace{-0.8em}
\label{tab:gsm8k}
\end{table}


\subsection{Mathematical Reasoning}
In this task, we evaluate our framework on GSM8k~\citep{cobbe2021training}, a collection of grade school math problems that take two to eight steps to solve. Models are expected to carry out step-by-step reasoning and arrive at the answer. Following \citet{lmsf}, a group of task instructions is adopted to prompt the model to first reason about the math problem and then yield an integer as its answer. For JSON format output, we prompt the model to output valid JSON blob with keys \textit{``reason''} and \textit{``answer.''} For natural language output, format instructions are used to encourage the model to generate the template phrase \textit{``The final answer is ...''} Meanwhile, this phrase is specified as a key phrase to appear in \framework's generation. 

\paragraph{Evaluation Metrics.} We measure \textit{Format Compliance} as the rate to which the generated answer follows format requirement. 
In addition, we measure \textit{Accuracy} as exact match of ground truth. 

\paragraph{Results.} As shown in \Cref{tab:gsm8k}, \textit{Prompt-Only (NL)} offers decent performance, with \textit{Llama} scoring $82.3\%$, \textit{Qwen2.5} $83.6\%$, and \textit{Qwen3} $90.5\%$ on accuracy. However, unstructured generation methods completely rely on the LLM for following format constraints and thus suffer from low compliance rate. Structured generation (\textit{Outlines}), on the contrary, guarantees format compliance, but its invasive intervention compromises task performance. While both Ctrl-G and \framework guarantee format compliance, \framework benefits from more accurate control signal provided by the instruction-aware HMM and achieves the best performance across all three models. 
In practice, we observe that \textit{Qwen} models have more skewed token distribution. We thus raise the control factor $\gamma$ to exert stronger control on the output.

\subsection{Event Argument Extraction}
The generative event argument extraction (EAE) task assesses a model's ability in identifying role-related arguments from source text. We evaluate on the ACE05-EN dataset~\citep{doddington2004automatic}, in which a model is presented with an article, a trigger word, and a set of roles to determine whether arguments associated with the roles are present in the article. This is naturally a templated task as generative models have to specify which word is extracted for which role. Regarding JSON output, we ask model to generate a JSON blob with roles as keys and extracted arguments as values. For natural language output, we specify the template \textit{``The <role$_i$> is ...''} for every relevant role. For \framework, we construct a flexible DFA that fuses the template phrases together with empty slots allowing LLM predict arguments spanning from 1 to 5 tokens.

\paragraph{Evaluation Metrics.} We measure performance by calculating the \textit{f1-score} comparing the extracted tuples and the ground truth tuples for the following categories:
Argument Id (AI), Argument Class (AC), Argument-attached Id (AI+), Argument-attached Class (AC+). 



\paragraph{Results.} The \textit{f1-scores} reported in \Cref{tab:ace05} suggest that EAE remains a challenging task for generative models. LLMs suffer from identifying correct relations in the article and presenting valid predictions that exist in the original text. Baseline methods show inconsistent trends across models, indicating LLMs' lack of robustness in event argument extraction. 
Employing \framework enhances overall extraction quality for \textit{Llama} and \textit{Qwen3}, while mainly improving over AI and AI+ for \textit{Qwen2.5}.
\framework's gain on AI and AI+ is more evident than its improvement on AC and AC+, suggesting that \framework can further benefit from a tighter association between roles and extracted arguments.

\definecolor{modelrowcolor}{gray}{0.90}

\begin{table}[htbp]
\centering
\small
\setlength{\tabcolsep}{2pt}
\renewcommand{\arraystretch}{1.1}
\begin{tabular}{l l c c c c}
\toprule
\textbf{Model} & \textbf{Method} & \textbf{AI} & \textbf{AC} & \textbf{AI+} & \textbf{AC+} \\
\midrule

\multirow{6}{*}{\shortstack[l]{\textit{Llama-3.1-8B}\\\textit{-Instruct}}} 
 & Prompt-Only (NL) & 36.8 & 27.3 & 34.8 & 25.5 \\
 & Prompt-Only (JSON) & 35.2 & 26.2 & 33.4 & 24.6 \\
 & Outlines (NL) & 37.1 & 27.8 & 35.1 & 26.0 \\
 & Outlines (JSON) & 33.6 & 25.2 & 31.6 & 23.6 \\
 & Ctrl-G & 39.1 & 27.5 & \textbf{37.0} & 26.1 \\
 & \textbf{\framework} & \textbf{39.4} & \textbf{28.7} & \textbf{37.0} & \textbf{26.8} \\

\midrule

\multirow{6}{*}{\shortstack[l]{\textit{Qwen2.5-7B}\\\textit{-Instruct}}} 
 & Prompt-Only (NL) & 32.6 & 25.5 & 31.2 & 24.4 \\
 & Prompt-Only (JSON) & 31.9 & 24.1 & 30.5 & 22.9 \\
 & Outlines (NL) & 33.2 & 24.9 & 31.1 & 23.7 \\
 & Outlines (JSON) & 34.1 & \textbf{26.1} & 32.5 & \textbf{24.7} \\
 & Ctrl-G & 34.7 & 24.5 & 33.1 & 23.9 \\
 & \textbf{\framework}$^{\gamma=2}$ & \textbf{35.2} & 25.9 & \textbf{33.4} & 24.5 \\

\midrule

\multirow{6}{*}{\textit{Qwen3-8B}} 
 & Prompt-Only (NL) & 33.2 & \textbf{24.6} & 31.5 & \textbf{23.1} \\
 & Prompt-Only (JSON) & 31.7 & 23.4 & 30.1 & 21.8 \\
 & Outlines (NL) & 33.3 & 24.2 & 31.5 & 22.6 \\
 & Outlines (JSON) & 32.5 & 23.0 & 30.8 & 21.5 \\
 & Ctrl-G & 33.7 & 23.3 & 31.6 & 22.4 \\
 & \textbf{\framework}$^{\gamma=2}$ & \textbf{34.0} & \textbf{24.6} & \textbf{32.1} & \textbf{23.1} \\

\bottomrule
\end{tabular}
\caption{Generative EAE results on ACE05.}
\vspace{-0.8em}
\label{tab:ace05}
\end{table}

\definecolor{modelrowcolor}{gray}{0.90}      
\definecolor{avgcolumncolor}{HTML}{EAEAEA} 

\renewcommand{\thefootnote}{\fnsymbol{footnote}}

\begin{table*}[t]
\centering
\small
\setlength{\tabcolsep}{3.5pt} 
\renewcommand{\arraystretch}{1.1}

\definecolor{avgcolumncolor}{gray}{0.95}

\begin{tabular}{l l c c c c c c c c >{\columncolor{avgcolumncolor}}c >{\columncolor{avgcolumncolor}}c >{\columncolor{avgcolumncolor}}c}
\toprule
\multirow{2}{*}{\textbf{Model}} & \multirow{2}{*}{\textbf{Method}} & \multicolumn{2}{c}{\textbf{Coherence}} & \multicolumn{2}{c}{\textbf{Consistency}} & \multicolumn{2}{c}{\textbf{Fluency}} & \multicolumn{2}{c}{\textbf{Relevance}} & \multicolumn{3}{c}{\cellcolor{avgcolumncolor}\textbf{Avg}} \\
\cmidrule(lr){3-4} \cmidrule(lr){5-6} \cmidrule(lr){7-8} \cmidrule(lr){9-10} \cmidrule(lr){11-13}
& & \textbf{$\rho$} & \textbf{$\tau$} & \textbf{$\rho$} & \textbf{$\tau$} & \textbf{$\rho$} & \textbf{$\tau$} & \textbf{$\rho$} & \textbf{$\tau$} & Format & \textbf{$\rho$} & \textbf{$\tau$} \\
\midrule

\multirow{6}{*}{\shortstack[l]{\textit{Llama-3.1-8B}\\\textit{-Instruct}}} 
 & Prompt-Only (NL) & 0.381 & 0.311 & 0.383 & 0.351 & 0.321 & 0.291 & 0.405 & 0.337 & 95.8 & 0.372 & 0.322 \\
 & Prompt-Only (JSON) & 0.449 & 0.368 & 0.446 & 0.415 & 0.326 & 0.296 & 0.424 & 0.358 & 99.8 & 0.411 & 0.359 \\
 & Outlines (NL) & 0.376 & 0.308 & 0.375 & 0.343 & 0.316 & 0.287 & 0.435 & 0.364 & 100 & 0.376 & 0.325 \\
 & Outlines (JSON) & 0.450 & 0.369 & \textbf{0.447} & \textbf{0.416} & \textbf{0.334} & \textbf{0.302} & 0.424 & 0.358 & 100 & 0.414 & 0.361 \\
 & Ctrl-G & 0.449 & 0.368 & 0.440 & 0.403 & 0.327 & 0.296 & 0.437 & 0.369 & 100 & 0.413 & 0.359 \\
 & \textbf{\framework} & \textbf{0.458} & \textbf{0.379} & 0.439 & 0.404 & 0.331 & 0.298 & \textbf{0.441} & \textbf{0.371} & 100 & \textbf{0.418} & \textbf{0.363} \\

\midrule

\multirow{6}{*}{\shortstack[l]{\textit{Qwen2.5-7B}\\\textit{-Instruct}}} 
 & Prompt-Only (NL) & 0.407 & 0.339 & 0.442 & 0.407 & 0.291 & 0.265 & 0.399 & 0.340 & 100 & 0.385 & 0.338 \\
 & Prompt-Only (JSON) & 0.411 & 0.334 & 0.488 & 0.455 & 0.305 & 0.280 & 0.383 & 0.326 & 99.7 & 0.396 & 0.349 \\
 & Outlines (NL) & 0.403 & 0.337 & 0.448 & 0.412 & 0.279 & 0.254 & 0.408 & 0.347 & 100 & 0.384 & 0.338 \\
 & Outlines (JSON) & \textbf{0.412} & \textbf{0.335} & 0.489 & 0.457 & 0.309 & 0.284 & 0.387 & 0.330 & 100 & 0.399 & 0.351 \\
 & Ctrl-G & 0.347 & 0.286 & 0.481 & 0.450 & 0.325 & 0.293 & 0.441 & 0.372 & 100 & 0.399 & 0.350 \\
 & \textbf{\framework}$^{\gamma=2}$ & 0.327 & 0.271 & \textbf{0.506} & \textbf{0.470} & \textbf{0.348} & \textbf{0.311} & \textbf{0.452} & \textbf{0.380} & 100 & \textbf{0.408} & \textbf{0.358} \\

\midrule

\multirow{6}{*}{\textit{Qwen3-8B}} 
 & Prompt-Only (NL) & \textbf{0.510} & \textbf{0.416} & 0.540 & 0.504 & 0.479 & 0.441 & 0.464 & 0.392 & 100 & 0.498 & 0.439 \\
 & Prompt-Only (JSON) & 0.504 & 0.409 & 0.491 & 0.459 & 0.444 & 0.410 & 0.450 & 0.382 & 99.8 & 0.472 & 0.415 \\
 & Outlines (NL) & 0.507 & 0.413 & 0.544 & 0.509 & 0.477 & 0.439 & 0.468 & 0.396 & 100 & 0.499 & 0.439 \\
 & Outlines (JSON) & 0.486 & 0.393 & 0.486 & 0.454 & 0.406 & 0.376 & 0.441 & 0.374 & 100 & 0.455 & 0.399 \\
 & Ctrl-G & 0.484 & 0.391 & \textbf{0.550} & \textbf{0.519} & 0.476 & 0.436 & 0.491 & 0.412 & 100 & 0.500 & 0.440 \\
 & \textbf{\framework}$^{\gamma=2}$ & 0.490 & 0.395 & 0.546 & 0.516 & \textbf{0.499} & \textbf{0.456} & \textbf{0.494} & \textbf{0.414} & 100 & \textbf{0.507} & \textbf{0.445} \\

\bottomrule
\end{tabular}
\caption{SummEval results, measured over Coherence, Consistency, Fluency, and Relevance.}
\vspace{-0.8em}
\label{tab:summeval}
\end{table*}
\subsection{LLM-as-a-judge Evaluation}
\label{app:llm-as-a-judge}
We then use LLMs as judges to evaluate the quality of summaries and assess how well it aligns with human annotation.
This evaluation is performed on the SummEval~\citep{fabbri2021summeval} which consists 1600 machine-generated summaries for 100 news articles and human annotated scores over four dimensions: Coherence, Consistency, Fluency, and Relevance. The models are asked to analyze the summary and assign a score from 1 to 5 based on the given criteria suggested by ChatGPT~\citep{OpenAI_ChatGPT_Software_2025}. 
We use the format ``\textit{The rating is ...}'' for natural language output and ``\textit{rating}'' as the key for harnessing JSON output.

\paragraph{Evaluation Metrics.} Following \citet{liu2023g}, we adopt the summary-level Spearman and Kendall-Tau correlation to gauge the performance of each method. Higher number indicates better alignment with human annotated scores.

\paragraph{Results.} For this task, \textit{Qwen} models perform well in following the output format in unstructured settings, securing over $99.7\%$ compliance rate. This may attribute to a less intensive reasoning phase compared to mathematical reasoning. As indicated in \Cref{tab:summeval}, \framework demonstrates strongest average correlation with human annotator, enhancing over Consistency, Fluency, and Relevance when applied to \textit{Qwen} models. A qualitative inspection reveals that \framework enables the flexible integration of scoring phrases at arbitrary positions, whereas unstructured and structured baselines consistently append ratings only after the analysis concludes.
\section{Analysis}
\subsection{Complexity and Efficiency}
\label{sec:efficiency}

\begin{figure*}[t]
    \begin{subfigure}[b]{0.32\textwidth}
        \centering
        \includegraphics[width=\textwidth]{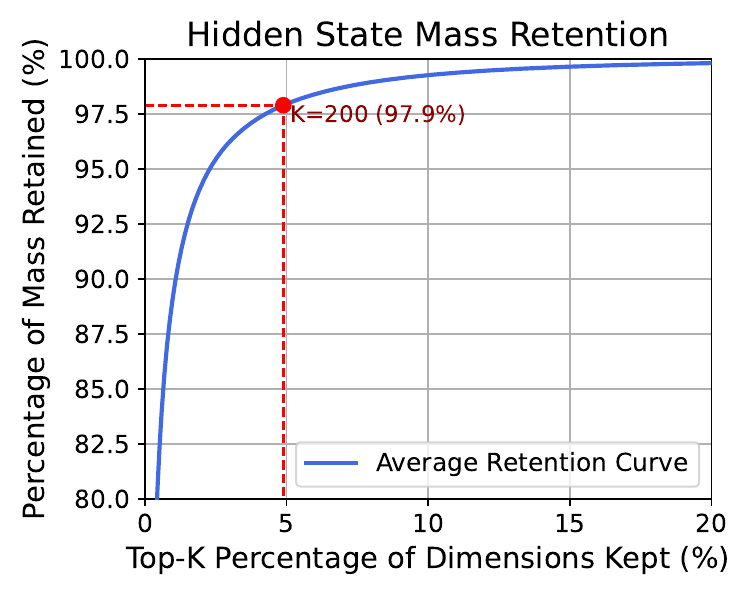}
        \caption{\textit{Llama-3.1-8B-Instruct}}
        \label{fig:first}
    \end{subfigure}
    \hfill 
    \begin{subfigure}[b]{0.32\textwidth}
        \centering
        \includegraphics[width=\textwidth]{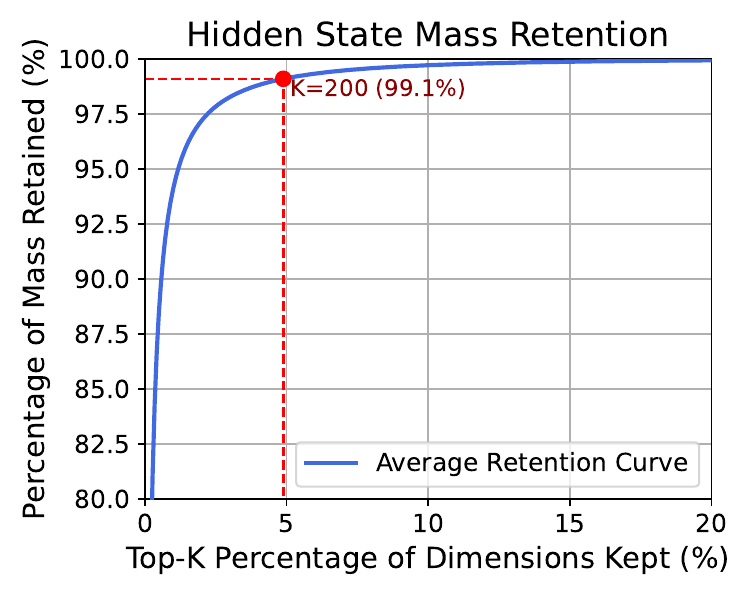}
        \caption{\textit{Qwen2.5-7B-Instruct}}
        \label{fig:second}
    \end{subfigure}
    \begin{subfigure}[b]{0.32\textwidth}
        \centering
        \includegraphics[width=\textwidth]{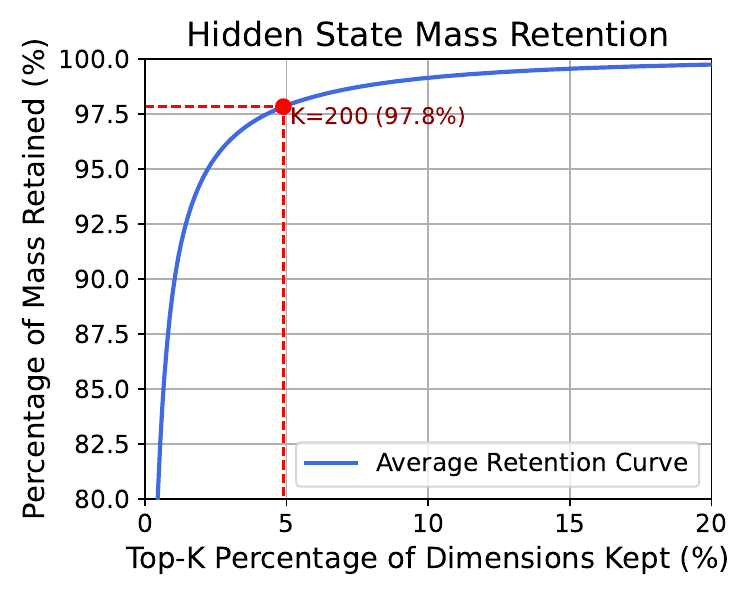}
        \caption{\textit{Qwen3-8B}}
        \label{fig:third}
    \end{subfigure}
    \caption{Average retention rate (of total mass) over $\topk$ HMM hidden states on GSM8k dataset.}
    \vspace{-0.5em}
    \label{fig:twofigs}
\label{fig:retention}
\end{figure*}

We analyze the efficiency of \framework from both theoretical complexity and empirical latency perspectives, highlighting improvements brought by our HMM pruning (Section~\ref{sec:pruning_methodology}) and flexible trie construction (Section~\ref{sec:trie}).

\definecolor{darkgreen}{rgb}{0.0, 0.5, 0.0}

\begin{table}[htbp]
\centering
\small 
\definecolor{modelrowcolor}{gray}{0.90} 
\vspace{0.5em}
    \begin{tabular}{l l}
    \toprule
    \textbf{Method} & \textbf{Acc (\%)} \\
    \midrule
    \multicolumn{2}{c}{\textit{Llama-3.1-8B-Instruct}} \\
    \cmidrule(lr){1-2}
    \framework w/o Pruning & 85.4 \\
    \framework & 85.2 \textcolor{red}{($\Delta$=-0.2)}\\
    \midrule
    \multicolumn{2}{c}{\textit{Qwen2.5-7B-Instruct}} \\
    \cmidrule(lr){1-2}
    \framework w/o Pruning & 89.8 \\
    \framework & 88.6 \textcolor{red}{($\Delta$=-1.2)}\\
    \midrule
    \multicolumn{2}{c}{\textit{Qwen3-8B}} \\
    \cmidrule(lr){1-2}
    \framework w/o Pruning & 90.8 \\
    \framework & 91.7 \textcolor{darkgreen}{($\Delta$=+0.9)}\\
    \bottomrule
    \end{tabular}
\caption{\framework performance on GSM8k with and without pruning.}
\vspace{-1em}
\label{tab:hmm}
\end{table}
\paragraph{Hidden State Sparsity.}
Our pruning strategy is empirically justified by the highly concentrated nature of the hidden state distributions (\Cref{fig:retention}). For \textit{Llama}- and \textit{Qwen}-distilled HMMs, the top 5\% of hidden states ($k=200$) capture over 97.8\% of the total probability mass on average.
Crucially, this approximation preserves task performance. As shown in \Cref{tab:hmm}, the accuracy degradation on GSM8k is minimal when pruning is applied (e.g., -0.2\% for \textit{Llama}, +0.9\% for \textit{Qwen3}). This confirms that the pruned HMM retains sufficient guidance information to steer generation effectively.

\paragraph{Computational Complexity.}
The primary bottleneck in HMM-guided generation is the emission probability calculation, which typically scales with $O(h|\mathcal{V}|)$. By exploiting the sparsity established above, we reduce this complexity to $O(k|\mathcal{V}|)$ where $k \ll h$. With $k=200$, this yields a $\sim13\times$ reduction in guidance FLOPs (from 1.08 down to 0.08 GFLOPs for an 8B model). Consequently, the total computational cost of the FEM constitutes only $\sim0.53\%$ of the LLM's main forward pass~\citep{kaplan2020scaling}, rendering the overhead negligible.

\paragraph{Empirical Latency.}
We measure the wall-clock latency of \framework against standard unconstrained generation, structured generation baselines (Outlines), and prior HMM-based control (Ctrl-G). As shown in Table~\ref{tab:latency}, \framework incurs minimal latency overhead while guaranteeing format compliance.

\begin{table}[h]
    \centering
    \small
    \setlength{\tabcolsep}{2pt}
    \renewcommand{\arraystretch}{1.2}
    \newcommand{\cmark}{\ding{51}}
    \newcommand{\xmark}{\ding{55}}
    
    \resizebox{\columnwidth}{!}{
    \begin{tabular}{l c c c c}
    \toprule
    \textbf{Method} & \textbf{Latency} & \textbf{Overhead} & \textbf{Format} & \textbf{Task Perf.} \\
    & \textit{(ms/token)} $\downarrow$ & \textit{(latency)} $\downarrow$ & & \\
    \midrule
    Prompt-Only & 3.07 & 1.00$\times$ &   \\
    Structured (Outlines) & 8.09 & 2.64$\times$ & \cmark &  \\
    Ctrl-G & 10.70 & 3.49$\times$ & \cmark &  \\
    \midrule
    \textit{\framework (Full HMM)} & 9.45 & 3.08$\times$ & \cmark & \cmark \\
    \textbf{\framework}  & \textbf{5.12} & \textbf{1.67}$\times$ & \cmark & \cmark \\
    \bottomrule
    \end{tabular}
    }
    \caption{Inference latency comparison on GSM8k with single NVIDIA A100.}
    \vspace{-1em}
    \label{tab:latency}
\end{table}

\paragraph{Scalability with Constraint Complexity.}
Beyond raw speed, \framework demonstrates superior scalability when handling complex constraints compared to baselines:
\begin{itemize}[topsep=0.1em, itemsep=0.1em, leftmargin=16pt]
    \item \textbf{vs. Structured Generation:} Frameworks like Outlines~\citep{outlines} and Guidance~\citep{guidance_ai_2024_guidance} rely on intersecting FSMs with the LLM vocabulary. This becomes computationally prohibitive for ``floating'' constraints at \textit{arbitrary positions} (e.g., after free-form reasoning), as modeling such unconstrained spans over long sequences triggers state-space explosion in the automaton.
    In contrast, \framework naturally handles these ``floating'' constraints, allowing the LLM to generate free-form text until the probabilistic guidance steers it into the target format, without requiring expensive pre-computation of all possible prefix permutations.

    \item \textbf{vs. Ctrl-G:} Ctrl-G \citep{ctrlg} fails on templates with chained variable-length segments, such as EAE extraction slots (e.g., ``\texttt{Role A: [span] Role B: [span]}''). Standard DFA construction creates unique paths for every length permutation, causing exponential state explosion ($O(K^N)$ for $N$ slots). Our \textit{Flexible Trie} (Section~\ref{sec:trie}) avoids this by merging wildcard paths at each fixed pivot, achieving linear complexity ($O(N \times K)$).
\end{itemize}

\subsection{The Steering Process}
\framework takes advantage of HMM to estimate the future format satisfaction rate and adjust token probabilities based on the estimation. To better understand this steering process, we examine the control signals produced by the FEM and visualize the control for a span of decoding step. We track the original LLM distribution, FEM distribution, and their composed distribution, which \framework decodes from. \Cref{fig:visualization} provides an illustration of \framework encouraging the generation of key phrase after step by step reasoning. While the LLM tends to conclude its response with ``\textit{The total number of ... is ...}'' \framework assigns high probabilities to the token ``\textit{final},'' steering the LLM generation to conform with format constraints. 

As LLMs are trained to provide clear and concise response, they tend to avoid repeating themselves when presenting the final answer. \framework captures this intricacy and replaces LLM's intended conclusive phrase with the format phrase ``\textit{The final answer is}'' to reduce repetition. We consider this format integration to be more natural than forcing LLM to generate certain phrases as in regex-structured generation.

\section{Related work}
In the paper, we explore a controllable text generation (CTG) method to decouple task solving from format adherence. There are two branches in CTG that provide avenues for achieving this format-task decoupling---content-wise hard control and attribute-wise soft control.
\textbf{Content-wise methods}~\citep{outlines, guidance_ai_2024_guidance} enforce strict adherence to predefined templates, ensuring reliability but often exerting invasive control that leads to abrupt, incoherent cut-offs.
\textbf{Attribute-wise soft control} offers greater flexibility, either by updating model weights---via retraining~\citep{keskar2019ctrl, arora2022director}, fine-tuning~\citep{wei2021finetuned, zeldes2020technical, li2021prefix, lester2021power}, or reinforcement learning~\citep{ouyang2022training, stiennon2020learning, zeng2024token, daisafe}---or through \textit{weighted decoding}~\citep{dathathri2019plug, yang2021fudge, krause2021gedi, schick2021self, liu2021dexperts, khandelwalgeneralization, sitdikov2022classifiers, wen2023grace, deng2023reward, LoulaLDLPG0EFEC25, lipkin2025fastcontrolledgenerationlanguage}. The latter steers generation at inference time using lightweight auxiliary models (Bayes' rule) without the high cost of retraining.
\framework adopts this weighted decoding paradigm to compute the posterior probability of format constraints as a decoupled attribute.

\section{Conclusion}
We present \framework, a novel decoding framework designed to decouple the responsibilities of task reasoning and format adherence. It achieves this responsibility separation by employing an auxiliary Format Estimation Module
to estimate future format satisfaction and modify token probabilities, thus allowing the LLM to concentrate solely on problem-solving. Experiments on mathematical reasoning, event argument extraction, and LLM-as-a-judge evaluation demonstrate this decoupling approach leads to overall performance gain, attributed to improved format compliance and more natural format integration in the response. 
\section*{Limitations}
\label{app:limitation}
While we show \framework enhances LLM task performance in various tasks, a few limitations should be taken into consideration when using \framework. Firstly, the HMM used to estimate format satisfaction rate is specific to an LLM, meaning that one has to distill a new HMM when switching to a different LLM. Although, in practice, we find that an HMM can be applied to larger LLMs in the same family, it is not an accurate representation of their token distributions.
Secondly, similar to other CTG methods that includes additional module for attribute modeling, \framework introduces additional computation overhead during decoding. As suggested by \citep{ctrlg}, the probabilistic traversal of future generation courses using HMM has a complexity that is linear to the number of edges in DFA and quadratic to the number of hidden states in HMM. Complex format constraints, converted to larger DFA, are thus likely to increase generation runtime. Finally, finding the optimal hyperparameter $\gamma$ for LLMs with highly peaked token distributions may require empirical explorations. Such distributions require increased $\gamma$ to ensure robust format compliance, yet an excessive value may adversely affect the output quality.
\section*{Ethical considerations}
We conduct experiments on mathematical reasoning, event argument extraction, and LLM-as-a-judge evaluation. 
The score assigned by an LLM should not be considered an accurate reflection of quality of the summary. In addition, the LLM responses to the GSM8k questions should not be referenced for math instruction as they may include hallucination.

We acknowledge the use of AI assistants for improving the manuscript's prose, generating tables in LaTeX format, figure design, and assisting with code implementation for the analysis of HMM hidden state pruning. All generated content, particularly the data in tables, was manually verified for accuracy against our experimental results.


\bibliography{custom}

\clearpage
\newpage
\appendix
\section{Probabilistic Reasoning over Logical Constraints}
\label{app:probabilistic}
\citet{gelato} and \citet{ctrlg} use a TPM to perform a probabilistic lookahead---that is, to efficiently compute $P_{\mathrm{TPM}}(\alpha|x_{\leq t})$, the probability that the full generated sequence will satisfy the constraint $\alpha$. The constraint is encoded as a DFA, and compliance is denoted by the event $\mathcal{D}(\alpha) = 1$. Then, the marginal probability over all sequences accepted by $\mathcal{D(\alpha)}$ is expressed as
\begin{equation*}
 P_\mathrm{TPM}(\alpha|x_{\leq t})=\frac{P_\mathrm{TPM}(\mathcal{D}(\alpha)=1, x_{\leq t})}{P_\mathrm{HMM}(x_{\leq t})}
\end{equation*}
where the numerator---likelihood of satisfying the constraint given a prefix $x_{\le t}$---is found by marginalizing over the HMM hidden states $z_t$ and the DFA states $s_t$
\begin{equation*}
\begin{split}
& P_{\mathrm{TPM}}(\mathcal{D}(\alpha)=1, x_{\le t}) = \\
& \quad \sum_{z_t} P_{\mathrm{TPM}}(\mathcal{D}(\alpha)=1 \mid z_t, s_t) P_{\mathrm{HMM}}(z_t, x_{\le t})
\end{split}
\end{equation*}
The conditional compliance probability $P_{\mathrm{TPM}}(\mathcal{D}(\alpha)=1 | z_t, s_t)$, as shown in~\citet{ctrlg}, is calculated using a backward recurrence relation. This sums the probabilities of all valid transitions from step $t$ to $t+1$, weighted by the HMM’s transition and emission probabilities
\begin{equation*}
\begin{split}
& P(\mathcal{D}(\alpha)=1 \mid z_t, s_t) = \\
& \quad \sum_{z_{t+1}} P(z_{t+1} \mid z_t) \sum_{s_{t+1}} P(\mathcal{D}(\alpha)=1 \mid z_{t+1}, s_{t+1}) \\
& \quad \sum _{\delta(s_t, x_{t+1})=s_{t+1}} P(x_{t+1} \mid z_{t+1})
\end{split}
\end{equation*}
The computed probability then serves as the tractable approximation of the format compliance likelihood, which is used to guide the LLM's next-token generation as shown in Equation \ref{eq:steering}.

\section{Suboptimal Control from Unconditioned HMM Distillation}
\label{app:unconditioned}
When an instruction-tuned model is prompted with no specific user input for unconditional sampling, it often defaults to generic conversational phrases like, ``\textit{Is there something I can help you with?}'' This behavior is a byproduct of its safety and helpfulness training. An HMM distilled from corpora containing such non-substantive responses learns a token distribution that is unrepresentative of the model's capabilities in actual problem-solving scenarios.

Consequently, when this HMM is applied to a complex reasoning task, it produce suboptimal control signal that can disrupt the reasoning process of the LLM. For example, with \textit{Llama}, when we apply Ctrl-G to the GSM8k dataset, its control mechanism prematurely forces the model to generate the required format phrase (``\textit{The final answer is ...}''), suppressing the step-by-step reasoning necessary to solve the problem. This results in an accuracy of 61.6\%, even worse than the Prompt-Only (NL) baseline.

Table \ref{tab:ctrlg-failure} provides an example of this failure mode on a GSM8k problem, contrasting Ctrl-G's flawed output with the coherent response from \framework, which uses an instruction-aware HMM.

\begin{table*}[h!]
\label{tab:ctrlg_vs_decog}
\small
\begin{tabularx}{\textwidth}{@{}lX@{}}
\toprule
\textbf{Method} & \textbf{Output} \\
\midrule
\textbf{Ctrl-G} & The final answer is 0.36 \textbf{(INCORRECT)}\\ 
\addlinespace[2ex] 
\textbf{\framework} & To find the probability that both tickets are winners, we need to multiply the probabilities of each ticket winning. \vspace{1ex}
    \begin{enumerate}
        \item The probability of the first ticket winning is 20\% or 0.2.
        \item The probability of the second ticket winning is three times more likely, so it's $3 \times 0.2 = 0.6$.
        \item The probability of both tickets winning is the product of their individual probabilities: $0.2 \times 0.6 = 0.12$.
        \item To express this as a percentage, we multiply by 100: $0.12 \times 100 = 12\%$.
    \end{enumerate} \vspace{1ex}
    The final answer is 12. \textbf{(CORRECT)} \\
\bottomrule
\end{tabularx}
\caption{Example of Ctrl-G's failure mode on GSM8k reasoning task.}
\label{tab:ctrlg-failure}
\end{table*}

\section{Token Entropy and Steering Strength}
\begin{figure}[h]
    \centering
    \includegraphics[width=1\linewidth, trim={0cm 0cm 1cm 0cm}, clip]{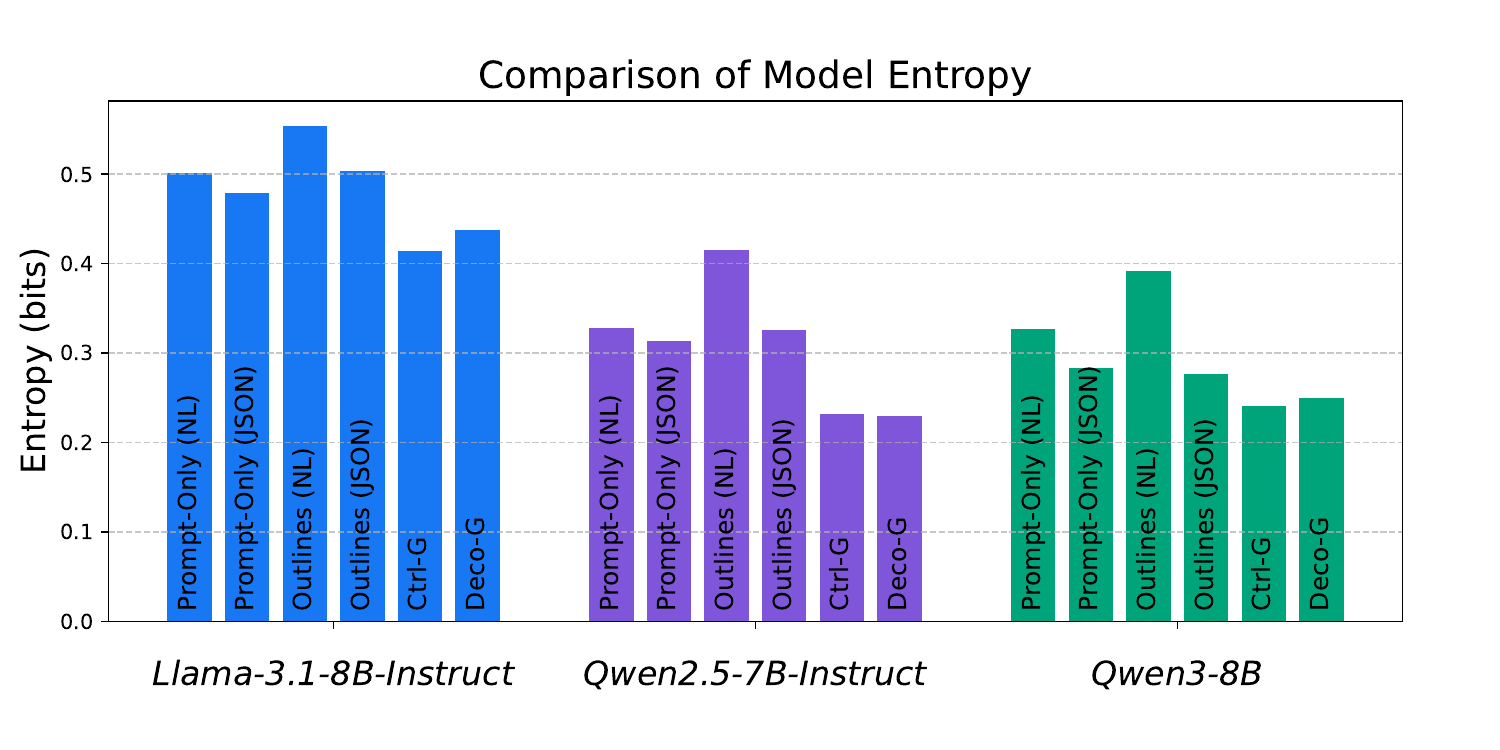}
    \caption{LLM's token-level entropy for different models and methods. \textit{Llama} has a more flexible token distribution as compared to \textit{Qwen}.}
    \label{fig:entropy}
\end{figure}

In the experiments, we report \framework's results with hyperparameter $\gamma=2$ for controlling \textit{Qwen} models, as $\gamma=1$ doesn't provide enough power to steer the model away from its own generation course. We hypothesize that \textit{Qwen} models' token distributions are more skewed than \textit{Llama}'s, making it difficult for the control signal to actually make an impact on the distribution. To verify this, we draw 100 examples from GSM8k responses and measure the average step-wise entropy of LLM token distribution. As shown in \Cref{fig:entropy}, \textit{Llama}'s entropy is significantly higher than those of \textit{Qwen2.5} and \textit{Qwen3}, suggesting that \textit{Llama}'s token probabilities are more spread out and diverse, whereas \textit{Qwen} models' token distributions are more peaky. This increased peakiness could be a consequence of the distribution squeezing induced by more intensive fine-tuning and preference optimization of the LLM \citep{learningdynamics}. It is thus intuitive to amplify \framework's control strength for LLM with more skewed distribution to guarantee format compliance.

Within the same model, structured generation methods have slightly higher entropy than their unstructured counterparts. This may attribute to imposed template tokens provoking more uncertainty in future token prediction. Meanwhile, \framework produces lowest LLM entropy, indicating that an absence of format constraint in task solving may lead to LLM providing the most confident response.

\begin{figure}[h]
    \centering
    \includegraphics[width=1\linewidth, trim={0.5cm 0cm 0cm 0cm}, clip]{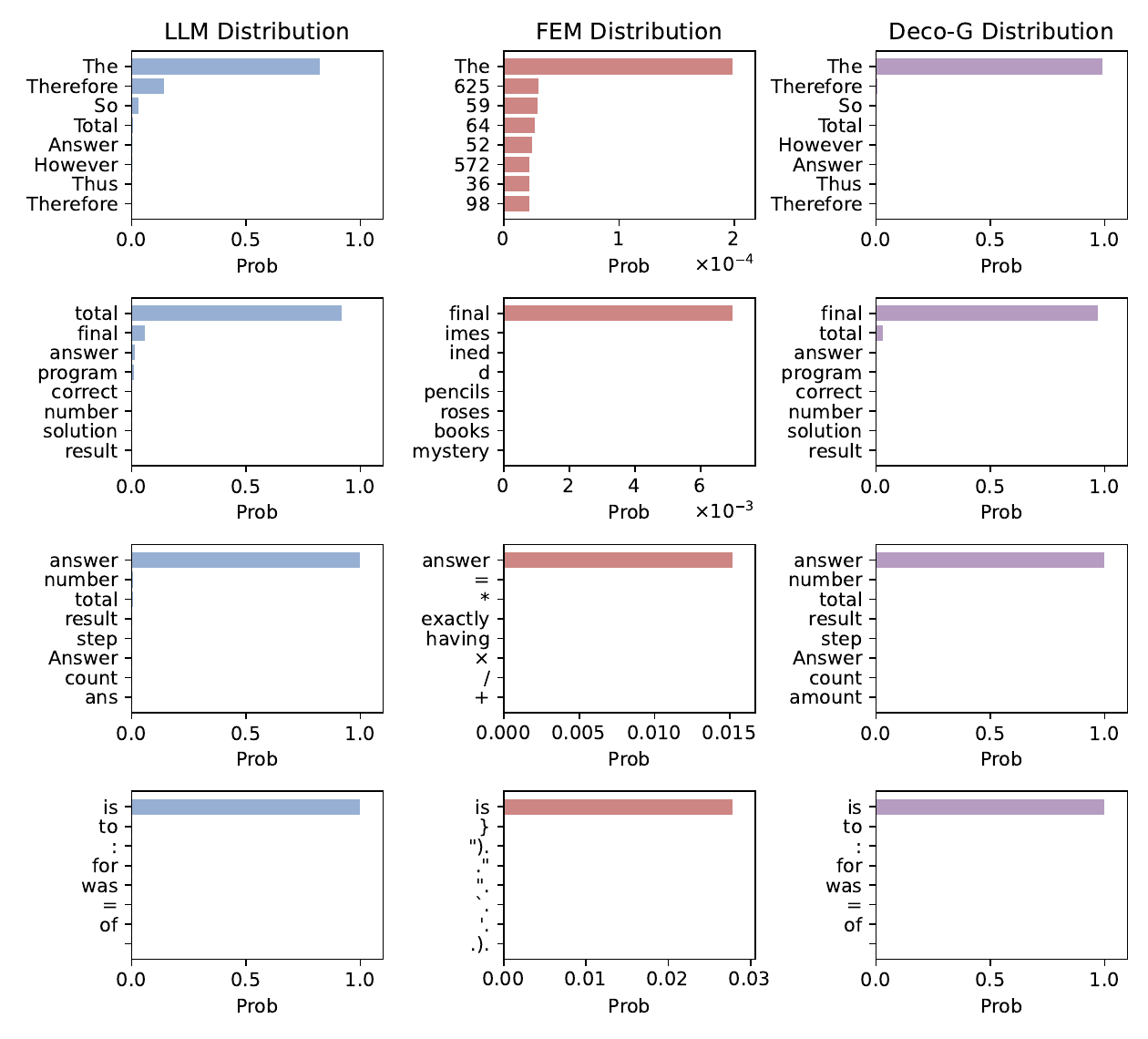}
    \caption{\framework steers \textit{Llama} to generate predefined template ``The final answer is ...'' by boosting probabilities of template tokens.
    }
    \label{fig:visualization}
\end{figure}

\section{HMM Pruning and Efficiency}
The primary benefit of HMM hidden state pruning is a substantial improvement in computational efficiency. By reducing the computation of the HMM emission stage, pruning achieves a 13x reduction in the FLOPs required by the HMM forward function at each decoding step (from approx. 1.08 GFLOPs to 0.08 GFLOPs for \textit{Llama}, calculation see below). When compared to the LLM's own forward pass, which requires approximately 16 GFLOPs per token~\citep{kaplan2020scaling}, the pruned FEM's computational cost constitutes only about 0.53\% of the main inference workload. This optimization renders the guidance overhead practically insignificant, thereby enhancing the viability of the DECO-G framework.

\paragraph{HMM Forward Computation Cost.}
\label{app:flops}
Below, we specify the computational cost (in FLOPs) of the HMM's forward pass. The calculation uses the HMM parameters for the \textit{Llama} model: hidden states $h$=4096, vocabulary size $|\mathcal{V}|$=128k, and $\topk$ states for pruning $k$=200.
\paragraph{Before Pruning.}
The total cost is the sum of the state transition cost ($2h^2$) and the emission cost ($2h|\mathcal{V}|$).
\begin{equation*}
\begin{split}
\text{Total FLOPs} &= (2 \times 4096^2) \\
&\quad + (2 \times 4096 \times 128{,}000) \\
&= (3.36 \times 10^7) + (1.05 \times 10^9) \\
&\approx \textbf{1.08 GFLOPs}
\end{split}
\end{equation*}

\paragraph{After Pruning.}
The cost is the sum of the state transition cost and the pruned emission cost ($2k|\mathcal{V}|$).
\begin{equation*}
\begin{split}
\text{Total FLOPs} &= (2 \times 4096^2) \\
&\quad + (2 \times 200 \times 128{,}000) \\
&= (3.36 \times 10^7) + (5.12 \times 10^7) \\
&\approx \textbf{0.08 GFLOPs}
\end{split}
\end{equation*}
This optimization reduces the HMM's computational overhead from 1.08 GFLOPs to 0.08 GFLOPs, a \textbf{$\sim$13x reduction} per decoding step.

\section{Reproducibility Statement}
A detailed description of our experimental setup, including the specific models used, HMM training parameters, and decoding strategy, is provided in the introductory paragraph of \Cref{sec:experiment}. To allow for replication of our experiments, the full prompts used for the mathematical reasoning (GSM8k), event argument extraction (ACE05), and LLM-as-a-judge (SummEval) tasks are detailed in \Cref{app:prompt}. Regarding computational overhead, a breakdown of the FLOPs required for the HMM forward pass, both with and without pruning, is presented in \Cref{app:flops}. 

\section{Prompt Construction}
\label{app:prompt}

We present the set of prompts used in the experiments. 
For GSM8k (\Cref{tab:GSM8k_Prompt}), we sample from a set of task instructions and a set of format instructions to construct prompts for baseline methods.
For ACE05 (\Cref{tab:ACE05_Prompt}), an event description is appended to the task instructions which further explains the event of interest.
For SummEval (\Cref{tab:SummEval_Prompt}), we include domain specific scoring criteria in the task instructions to help LLM align better with human annotations for all methods.

\begin{table*}[htbp]
\centering
\resizebox{\textwidth}{!}{%
\begin{tabular}{@{}llccccccccccccc@{}}
\toprule
\multirow{2}{*}{\textbf{Model}} & \multirow{2}{*}{\textbf{Method}} & \multicolumn{3}{c}{\textbf{AI}} & \multicolumn{3}{c}{\textbf{AC}} & \multicolumn{3}{c}{\textbf{AI+}} & \multicolumn{3}{c}{\textbf{AC+}} \\
\cmidrule(lr){3-5} \cmidrule(lr){6-8} \cmidrule(lr){9-11} \cmidrule(lr){12-14}
 &  & Precision & Recall & f1 & Precision & Recall & f1 & Precision & Recall & f1 & Precision & Recall & f1 \\ \midrule
\multirow{6}{*}{Llama-3.1-8B-Instruct} & Prompt-Only (NL) & 33.7 & 40.5 & 36.8 & 24.2 & 31.4 & 27.3 & 30.5 & 40.5 & 34.8 & 21.7 & 31.0 & 25.5 \\
 & Prompt-Only (JSON) & 30.5 & 41.8 & 35.2 & 21.5 & 33.3 & 26.2 & 27.8 & 41.7 & 33.4 & 19.6 & 33.0 & 24.6 \\
 & Outlines (NL) & 33.7 & 41.3 & 37.1 & 24.4 & 32.3 & 27.8 & 30.5 & 41.3 & 35.1 & 21.8 & 32.1 & 26.0 \\
 & Outlines (JSON) & 28.0 & 42.0 & 33.6 & 19.8 & 34.4 & 25.2 & 25.4 & 41.6 & 31.6 & 18.0 & 34.1 & 23.6 \\
 & Ctrl-G & 38.3 & 39.9 & 39.1 & 25.6 & 29.8 & 27.5 & 34.7 & 39.6 & 37.0 & 23.4 & 29.5 & 26.1 \\
 & \textbf{\framework} & 39.4 & 39.3 & 39.4 & 27.3 & 30.3 & 28.7 & 35.5 & 38.5 & 37.0 & 24.5 & 29.6 & 26.8 \\ \midrule
\multirow{6}{*}{Qwen2.5-7B-Instruct} & Prompt-Only (NL) & 29.7 & 36.2 & 32.6 & 22.6 & 29.4 & 25.5 & 27.5 & 35.7 & 31.2 & 21.0 & 29.0 & 24.4 \\
 & Prompt-Only (JSON) & 29.1 & 35.2 & 31.9 & 21.3 & 27.7 & 24.1 & 27.0 & 35.0 & 30.5 & 19.7 & 27.4 & 22.9 \\
 & Outlines (NL) & 28.6 & 39.5 & 33.2 & 20.7 & 31.4 & 24.9 & 26.2 & 39.0 & 31.3 & 19.1 & 31.2 & 23.7 \\
 & Outlines (JSON) & 32.2 & 36.2 & 34.1 & 23.9 & 28.7 & 26.1 & 29.6 & 36.1 & 32.5 & 21.9 & 28.4 & 24.7 \\
 & Ctrl-G & 29.9 & 41.3 & 34.7 & 20.1 & 31.4 & 24.5 & 27.8 & 41.0 & 33.1 & 19.3 & 31.4 & 23.9 \\
 & \textbf{\framework$^{\gamma = 2}$} & 30.9 & 41.0 & 35.2 & 21.8 & 32.0 & 25.9 & 28.4 & 40.5 & 33.4 & 20.1 & 31.4 & 24.5 \\ \midrule
\multirow{6}{*}{Qwen3-8B} & Prompt-Only (NL) & 28.0 & 40.9 & 33.2 & 19.5 & 33.3 & 24.6 & 25.6 & 40.9 & 31.5 & 17.8 & 33.0 & 23.1 \\
 & Prompt-Only (JSON) & 27.0 & 38.4 & 31.7 & 18.6 & 31.7 & 23.4 & 24.7 & 38.4 & 30.1 & 16.7 & 31.4 & 21.8 \\
 & Outlines (NL) & 27.9 & 41.4 & 33.3 & 18.9 & 33.6 & 24.2 & 25.4 & 41.3 & 31.5 & 17.1 & 33.2 & 22.6 \\
 & Outlines (JSON) & 26.8 & 41.4 & 32.5 & 17.5 & 33.4 & 23.0 & 24.5 & 41.3 & 30.8 & 15.9 & 33.2 & 21.5 \\
 & Ctrl-G & 27.6 & 43.2 & 33.7 & 17.5 & 35.0 & 23.3 & 24.8 & 43.5 & 31.6 & 16.5 & 34.8 & 22.4 \\
 & \textbf{\framework$^{\gamma = 2}$} & 27.9 & 43.6 & 34.0 & 18.9 & 35.1 & 24.6 & 25.5 & 43.4 & 32.1 & 17.3 & 34.8 & 23.1 \\ \bottomrule
\end{tabular}%
}
\caption{Full ACE05 Results.}
\label{tab:full_ace05}
\end{table*}
\section{More EAE Results}
In \Cref{tab:ace05}, we report the \textit{f1-scores} for each method. In \Cref{tab:full_ace05}, we present the full results for our event argument extraction experiment.

\section{HMM Distillation and Usage}
For \textit{Llama} and \textit{Qwen}, we distill their HMMs on the LLM continuation only, since the instructions from \textit{Natural-Instructions} are human authored and should not be considered reflecting LLM distribution. We remove the special chat tokens (e.g. <|system|>, <|user|>, etc.) from the responses for HMM to capture the natural language distribution. 

We tried different inputs to the HMM, including 1) regular prompt (with chat template), 2) cleaned text prompt (without chat template), and 3) no prompt (empty string). In practice, their results are almost identical. Nonetheless, in accordance with the distillation objective, we report scores yielded from using empty input to the HMM.


\begin{table*}[htb]
\centering
\small
\begin{tabular}{lp{9.5cm}}
\multicolumn{2}{l}{\textbf{GSM8k}}\\
\toprule
\textbf{Task Instructions} & 1. Follow the instruction to complete the task:\textbackslash nYou are a math tutor who helps students of all levels understand and solve mathematical problems. \textbackslash nRead the last question carefully and think step by step before answering, the final answer must be only a number.\\
\\
& 2. Follow the instruction to complete the task:\textbackslash nRead the last question carefully and think step by step before answering, the final answer must be only a number. You are a math tutor who helps students of all levels understand and solve mathematical problems.\\
\\
& 3. Follow the instruction to complete the task:\textbackslash nMathematical problem-solving task:\textbackslash n- Given: A mathematical question or problem\textbackslash n- Required: A numerical answer only\textbackslash n- Role: You are a math tutor assisting students of all levels\textbackslash n- Process: Think step by step to solve the problem\textbackslash nNote: Read the question carefully before beginning your analysis.\\
\midrule
\textbf{NL Format Instructions} & 
1. Provide your output in the following text format:\textbackslash n<think step by step>. The final answer is <answer>\\
\\
& 2. Provide your output in the following text format:\textbackslash nReasoning: <reasoning first>. Answer: The final answer is ...\\
\\

\textbf{JSON Format Instructions} & Provide your output in the following valid JSON format:\textbackslash n\{``reason'': ``<step by step reasoning>'',``answer": ``<final answer>''\}\\
\midrule
\textbf{Question Example} & Janet’s ducks lay 16 eggs per day. She eats three for breakfast every morning and bakes muffins for her friends every day with four. She sells the remainder at the farmers' market daily for \$2 per fresh duck egg. How much in dollars does she make every day at the farmers' market?
\\
\bottomrule
\end{tabular}
\caption{GSM8k prompt construction and an example question.} 
\label{tab:GSM8k_Prompt} 
\end{table*}

\begin{table*}[htb]
\centering
\small
\begin{tabular}{lp{9.5cm}}
\multicolumn{2}{l}{\textbf{ACE05}}\\
\toprule
\textbf{Task Instructions} & 
You are an argument extractor designed to check for the presence of arguments regarding specific roles for an event in a sentence.
\textbackslash nTask Description: Identify all arguments related to the role Attacker, Target, Instrument, Place, Agent in the sentence. These arguments should have the semantic role corresponding to the given event trigger by the word span between [t] and [/t].\\
\\
& The event of interest is Conflict:Attack. The event is related to conflict and some violent physical act. Roles of interest: Attacker, Target, Instrument, Place, Agent\\
\midrule
\textbf{NL Format Instructions} & 
Provide your output in the following text format:\textbackslash nThe <role\_1> is: <extracted argument>\textbackslash nThe <role\_2> is: <extracted argument>\textbackslash n...\textbackslash nThe <role\_n> is: <extracted argument>\\
\\
\textbf{JSON Format Instructions} &
Provide your output in the following valid JSON format:\textbackslash n\{``<role>'': ``<extracted argument>'' for role in roles of interest\}\\
\midrule
\textbf{Question Example} & Text: Efforts were to continue at the United Nations Friday to find a breakthrough in the diplomatic stalemate on Iraq , with Washington warning it could bypass the Security Council and go to [t] war [/t] alone .
\\
\bottomrule
\end{tabular}
\caption{ACE05 prompt construction and an example question.} 
\label{tab:ACE05_Prompt} 
\end{table*}

\begin{table*}[htb]
\vspace{-1em}
\centering
\small
\begin{tabular}{p{2.5cm} p{10.5cm}}
\multicolumn{2}{l}{\textbf{SummEval}}\\
\toprule
\textbf{Task Instructions (Coherence)} & 
You will be provided with a summary written for a news article
\textbackslash nYour task is to rate the summary based on its coherence.
\textbackslash n\textbackslash nPlease ensure you read and understand these instructions carefully. Keep this document open while reviewing, and refer to it as needed.
\textbackslash n\textbackslash nEvaluation Criteria:
\textbackslash nCoherence (1-5):
\textbackslash n- 5: The summary is well-structured and organized, presenting information in a logical and seamless flow.
\textbackslash n- 4: The summary is mostly coherent, with minor lapses in organization or flow.
\textbackslash n- 3: The summary has noticeable organizational issues or lacks a smooth flow but is somewhat understandable.
\textbackslash n- 2: The summary is poorly structured, with significant difficulties in following its logic or flow.
\textbackslash n- 1: The summary is highly disjointed and lacks any meaningful structure or coherence.
\textbackslash nUse these criteria to assign a coherence score between 1 and 5 based on how well the summary organizes and presents information in a clear and logical manner.
\\
\\
\textbf{Task Instructions (Consistency)} &
You will be provided with a news article and a summary written for this article.
\textbackslash nYour task is to rate the summary based on its consistency.
\textbackslash n\textbackslash nPlease ensure you read and understand these instructions carefully. Keep this document open while reviewing, and refer to it as needed.
\textbackslash n\textbackslash nEvaluation Criteria:
\textbackslash nConsistency (1-5):
\textbackslash n- 5: The summary is fully factually accurate and all its statements are directly supported by the source document.
\textbackslash n- 4: The summary is mostly factually accurate, with only minor errors or omissions.
\textbackslash n- 3: The summary contains noticeable factual errors or unsupported statements but retains some alignment with the source document.
\textbackslash n- 2: The summary has significant factual inaccuracies or includes multiple unsupported claims.
\textbackslash n- 1: The summary is largely inconsistent with the source, containing numerous factual inaccuracies or fabricated details.
\textbackslash nUse these criteria to assign a consistency score between 1 and 5 based on how well the summary aligns factually with the source article.
\\
\\
\textbf{Task Instructions (Fluency)} &
You will be provided with a summary written for a news article.
\textbackslash nYour task is to rate the summary based on its fluency.
\textbackslash n\textbackslash nPlease ensure you read and understand these instructions carefully. Keep this document open while reviewing, and refer to it as needed.
\textbackslash n\textbackslash nEvaluation Criteria:
\textbackslash nFluency (1-5):
\textbackslash n- 5: The summary is clear and easy to read, with good grammar, spelling, and sentence structure.
\textbackslash n- 4: The summary is generally clear and fluent, with a few minor errors that don't interfere with understanding.
\textbackslash n- 3: The summary has some noticeable issues that might make it a little harder to read but still understandable overall.
\textbackslash n- 2: The summary has more noticeable problems that might make it challenging to follow in places.
\textbackslash n- 1: The summary has significant errors that make it difficult to read or understand in many parts.
\textbackslash nUse these criteria to assign a fluency score between 1 and 5 based on the quality of grammar, word choice, and sentence structure.
\textbackslash nImportant: When evaluating fluency, ignore punctuation and capitalization. Focus only on how natural and easy the language feels regardless of formatting.
\\
\\
\textbf{Task Instructions (Relevance)} &
You will be provided with a summary written for a news article.
\textbackslash nYour task is to rate the summary based on its relevance.
\textbackslash n\textbackslash nPlease ensure you read and understand these instructions carefully. Keep this document open while reviewing, and refer to it as needed.
\textbackslash n\textbackslash nEvaluation Criteria:
\textbackslash nRelevance (1-5):
\textbackslash n- 5: The summary includes all the important information from the source document with no redundancies or irrelevant details.
\textbackslash n- 4: The summary is mostly relevant, with only minor omissions or slight redundancies.
\textbackslash n- 3: The summary includes some important information but misses key points or has noticeable redundancies.
\textbackslash n- 2: The summary contains limited relevant information, with significant omissions or excessive irrelevant content.
\textbackslash n- 1: The summary is largely irrelevant, failing to capture the main points of the source document.
\textbackslash nUse these criteria to assign a relevance score between 1 and 5 based on how well the summary captures the important content from the source without including excess or redundant information.
\\
\midrule
\textbf{NL Format Instructions} & 
Provide your output in the following text format: <analyze the summary>. The rating is <a number between 1 and 5>\\
\\
\textbf{JSON Format Instructions} & 
Provide your output in the following valid JSON format:\textbackslash n\{``analysis'': ``<analyze the summary>'',``rating'': <a number between 1 and 5>\}\\
\bottomrule
\end{tabular}
\caption{SummEval prompts.} 
\label{tab:SummEval_Prompt} 
\end{table*}

\end{document}